%% file: emnlp2022.tex
\pdfoutput=1

\documentclass[11pt]{article}

\usepackage{EMNLP2022}

\usepackage{times}
\usepackage{latexsym}

\usepackage[T1]{fontenc}

\usepackage[utf8]{inputenc}

\usepackage{microtype}

\usepackage{inconsolata}

\usepackage{url}            
\usepackage{booktabs}       
\usepackage{amsfonts}       
\usepackage{color,colortbl}
\usepackage{tabularx}
\usepackage{booktabs}
\usepackage{multirow}
\usepackage{multicol}
\usepackage{array}
\usepackage{xspace}
\usepackage{pifont}
\newcommand{\xmark}{\ding{55}}%
\newcommand{\longsysname}{Plug-and-Play VQA}
\newcommand{\sysname}{\textsc{PnP-VQA}}
\newcommand{\fullsysname}{\longsysname{} (\sysname)}
\usepackage{balance}
\usepackage{enumitem}
\usepackage{mathtools}
\usepackage{rotating}
\usepackage{tabulary}
\usepackage{subcaption}
\usepackage[export]{adjustbox}
\usepackage{ctable}
\usepackage{diagbox}
\usepackage{bbm}
\usepackage{makecell}
\usepackage[linesnumbered,ruled]{algorithm2e}
\usepackage[title]{appendix}
\usepackage{soul}

\graphicspath{{./figs/}}

\title{Plug-and-Play VQA: Zero-shot VQA by Conjoining Large Pretrained Models with Zero Training}

\author{Anthony Meng Huat Tiong$^{1,2}$, Junnan Li$^{1}$, Boyang Li$^{2}$, \\
	\textbf{Silvio Savarese}$^{1}$, and  \textbf{Steven C.H. Hoi}$^{1}$\\
	$^{1}$Salesforce Research \qquad $^{2}$Nanyang Technological University, Singapore\\
 { \texttt{\{anthony.tiong, junnan.li, ssavarese, shoi\}@salesforce.com}} \\
{\texttt{boyang.li@ntu.edu.sg}}\\
{\normalsize \url{https://github.com/salesforce/LAVIS/tree/main/projects/pnp-vqa}}}

\begin{document}

\maketitle
\input{sec_abstract}

\input{sec_introduction}

\input{sec_literature}

\input{sec_method}

\input{sec_experiment}
\input{sec_conclusion}

\bibliography{bib}
\bibliographystyle{acl_natbib}

\appendix

\input{sec_appendix}

\end{document}

%% file: sec_abstract.tex
\begin{abstract}

Visual question answering (VQA) is a hallmark of vision and language reasoning
and a challenging task under the zero-shot setting.
We propose \fullsysname,
a modular framework for zero-shot VQA.
In contrast to most existing works, which require substantial adaptation of pretrained language models (PLMs) for the vision modality,
\sysname{} requires no additional training of the PLMs.
Instead, we propose to use natural language and network interpretation as an intermediate representation that glues pretrained models together. We first generate question-guided informative image captions,
and pass the captions to a PLM as context for question answering.
Surpassing end-to-end trained baselines, \sysname{} achieves state-of-the-art results on zero-shot VQAv2~\cite{goyal2017making} and GQA~\cite{hudson2019gqa}. With 11B parameters, it outperforms the 80B-parameter Flamingo model~\cite{Deepmind:Flamingo2022} by 8.5\% on VQAv2. 
With 738M PLM parameters,
\sysname{} achieves an improvement of 9.1\% on GQA over FewVLM~\cite{fewvlm:jin2021good} with 740M PLM parameters.

\end{abstract}

%% file: sec_introduction.tex
\section{Introduction}
\label{sec:introduction}

Recent years have witnessed unprecedented performance gains on many natural language reasoning tasks, especially in zero-shot and few-shot settings, being derived from scaling up pretrained language models (PLMs) and their training data \cite{devlin2018bert,liu2019roberta,GPT-3,Raffel:JMLR:T5,gpt-neox-20b,sanh2022multitask:T0,Wei2021:FLAN}.  Inspired by their success, a natural thought is that utilizing PLMs should also boost zero-shot performance in vision-language reasoning tasks. 

However, to leverage PLMs for vision-language tasks, most existing methods require non-trivial adaptation of the PLMs for the vision modality, which necessitates the design of new network components and training objectives. For example, \citet{SungYiLin2022} and \citet{Deepmind:Flamingo2022} insert into the PLMs new layers that are trained from scratch. \citet{frozen:tsimpoukelli2021multimodal} train vision encoders that output soft prompts to frozen PLMs. \citet{chen2021visualgpt} and \citet{Eichenberg2021:MAGMA} train both the vision encoders and new layers inserted into PLMs. 
In the zero-shot setting, various vision-language pretraining objectives are employed,
such as image captioning~\cite{Deepmind:Flamingo2022} and image-conditioned masked language modeling~\cite{fewvlm:jin2021good}.

From the perspective of general-purpose AI, the ability to perform new tasks by simply recombining large-scale pretrained models, or foundation models \cite{foundation-models-2021}, without architectural changes or extra training would be highly desirable. Such a system would be able to dynamically adjust to previously unknown tasks by simply rewiring a small number of foundation models. However, to obtain high performance without some form of end-to-end training would seem difficult, if not impossible. 

We present \fullsysname{}, a framework for zero-shot visual question answering which conjoins large pretrained models with zero additional training and achieves state-of-the-art performance on zero-shot VQAv2~\cite{goyal2017making} and GQA~\cite{hudson2019gqa}. 
For the purpose of bridging the vision and language modalities, we employ a pretrained vision-language model (PVLM)~\cite{li2022blip} that describes visual information with textual captions. 
In order to obtain relevant and informative captions, we apply a network interpretability technique~\cite{selvaraju2017grad} to detect image patches that are relevant to the question. After that, we generate captions stochastically for these image patches. Finally, we employ a PLM~\cite{khashabi2022unifiedqa} to answer the question from the captions. 

Research in cognitive science and neuroscience suggests that the human cognitive system is largely modular~\cite{Shettleworth2012:Modularity,Bertolero2015:modular-brain}. For instance, the pioneering work of \citet{Fodor1983:modularity} argued that the low-level human cognition is constituted of several fast, autonomous, and domain-specific modules.
For purely practical purposes, a modular design of artificial general intelligence would make it easy to harness rapid progress in each individual component, as the components can be individually replaced and updated without affecting other parts of the system. 
With this paper, we offer such a modular design for zero-shot VQA that leverages recent advances in PLM and PVLMs and combines them with an innovative application of network interpretability.

\vspace{1ex}
\noindent We summarize our contributions as follows:
\begin{itemize}[leftmargin=*]
	\setlength\itemsep{0pt}
	\item We introduce \sysname{}, a modular framework for zero-shot VQA without training. Its flexibility allows \sysname{} to jointly evolve as pretrained models continue to advance.
	\item 
	Besides natural language, we propose the use of network interpretation as the interface between pretrained LMs and VLMs. With an interpretability technique, we create image captions that extensively cover information relevant to the question, which enable accurate QA. 
	\item We demonstrate state-of-the-art zero-shot VQA performance on multiple benchmarks. 
	On VQAv2, \sysname{}\textsubscript{11B} obtains 8.5\% improvement over Flamingo\textsubscript{80B}~\cite{Deepmind:Flamingo2022}, which applies extensive end-to-end VL-pretraining. On GQA, \sysname{}\textsubscript{large} outperforms FewVLM\textsubscript{large}~\cite{fewvlm:jin2021good} by 9.1\%. 
\end{itemize}

%% file: sec_literature.tex
\section{Related Work}
\label{sec:literature}
\input{table/fig_framework}
\noindent \textbf{Large-scale image-text pretraining} of neural networks is a popular research direction. 
Various vision-language pretraining tasks have been proposed, including image-conditioned language modeling~\cite{frozen:tsimpoukelli2021multimodal, Deepmind:Flamingo2022}, masked language modeling~\cite{tan2019lxmert, lu2019vilbert, li2020unimo}, prefix language modeling~\cite{wang2021simvlm},  image-text matching~\cite{li2019visualbert, chen2020uniter, li2020oscar} and image-text contrastive learning~\cite{radford2021learning, jia2021scaling, li2021align}.
After pretraining, several models exhibit zero-shot capabilities in image-text retrieval~\cite{jia2021scaling, radford2021learning, xvlm} and image captioning~\cite{wang2021simvlm, li2022blip}.
However, zero-shot VQA remains a challenging task due to its high requirement on the model's reasoning ability.

\vspace{0.1in}
\noindent \textbf{Adapting PLMs for zero-shot VQA} has shown promising results.
In order to incorporate vision information into PLMs,
most existing methods perform additional vision-language training on image-text data.
Frozen~\cite{frozen:tsimpoukelli2021multimodal} trains the vision encoder while keeping the gigantic PLM frozen to retain its knowledge in question answering. 
The output from the vision encoder is prepended to the text as prompts to the frozen language model.
FewVLM~\cite{fewvlm:jin2021good} finetunes the PLM using the prefix language modeling and masked language modeling objectives. VLKD~\cite{dai2022enabling} distills multimodal knowledge to PLM by using CLIP~\cite{radford2021learning} as the teacher model during finetuning. Flamingo~\cite{Deepmind:Flamingo2022} adds additional layers to both the pretrained vision model and the PLM and trains the new layers on billions of image-text pairs.

Different from the above work,
\sysname{} directly employs pretrained models with neither architectural modifications nor additional training.

Most similar to our work, PICa~\cite{pica:yang2021empirical} converts an image to a single caption and adopts GPT-3~\cite{GPT-3} for zero-shot VQA.
In comparison, \sysname{} generates multiple question-guided captions and performs fusion of captions after encoding to effectively utilize a large number of captions, yielding considerable performance gains. 

An orthogonal research direction for zero-shot VQA is to train the VLMs on synthetic VQA examples generated from captions~\cite{changpinyo2022all, banerjee2020weaqa}.
\sysname{} does not require additional training. 

\vspace{0.1in}
\noindent \textbf{Natural language as an intermediate representation} or interface between different models or multiple steps of reasoning is an emerging machine learning strategy. It dates back to at least \citet{andreas-etal-2018-Latent-Language} and saw renewed interest in the past few months due to the prevalence of large PLMs. \citet{andreas-etal-2018-Latent-Language} and \citet{vong2022fewshot} learn natural language descriptions that function as few-shot classifiers within an image-text matching model. \citet{Bostrom2022} generate intermediate reasoning steps with finetuned PLMs. \citet{Zhou-2022:Least-to-Most-Prompting} prompt a PLM to generate subproblem descriptions for a complex problem, and feed the subproblems back to the PLM to solve hierarchically. \citet{WuTongshuang2022} chain PLM outputs and inputs.  \citet{zeng2022socratic} show that language-conjoined LM and VLM successfully perform captioning and retrieval but do not evaluate their models on VQA. In comparison, \sysname{} adopts both natural language and network interpretation as the interface between different pretrained models.

%% file: table/fig_framework.tex
\begin{figure*}[!t]

	\centering
	\includegraphics[width=1\textwidth]{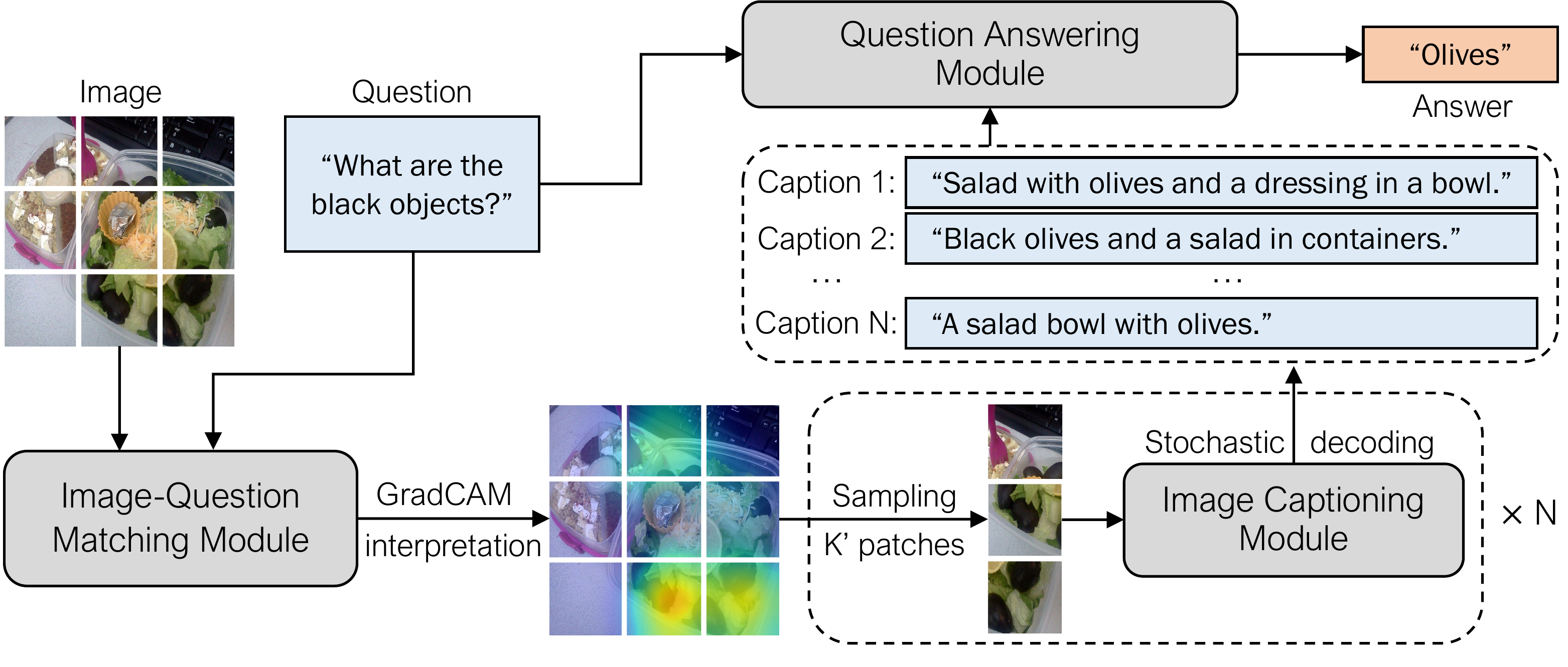}
	   \vspace{-3ex}
  \caption
  	{
  The system architecture of \sysname{},
  consisting of three pretrained modules:
  (1) an image-question matching module that identifies image patches relevant to the question, (2) an image captioning module that generates a diverse set of captions,
  (3) a question answering module that generates an answer given the question and captions. 
  For the image-question matching module and image captioning module, we adopt BLIP~\cite{li2022blip}. For the question answering module, we adopt UnifiedQAv2~\cite{khashabi2022unifiedqa}.  
	  } 
  \label{fig:framework}
 \end{figure*} 

%% file: sec_method.tex
\section{Method}
\label{sec:method}
The central idea of \fullsysname{} is to establish an interface between a pretrained language model and a pretrained vision-language model without training. We demonstrate that natural language image captions and network saliency maps together serve as an effective interface. 
Ideally, the generated captions should thoroughly cover information that is present in the image and be relevant to the question. We foster relevance by identifying image patches most related to the question with a saliency map-based interpretability technique and generating captions from these patches only. Further, we promote coverage by injecting stochasticity, including random sampling of relevant image patches and of the textual tokens during caption generation. 

The overall system architecture (Figure~\ref{fig:framework}) consists of three modules:
\begin{enumerate}
\itemsep0em
    \item an image-question matching module that identifies the relevant image patches given a question,
    \item an image captioning module that generates a diverse set of captions from a set of image patches, and
    \item a question answering module that outputs an answer given the question and the generated captions.
\end{enumerate}

In this section, we introduce the three modules in detail. 

\subsection{Matching Image Patches and Questions}
\label{sec:image-question-matching}
An image serves as a rich source of information, but the question at hand is likely focused only on particular objects or regions. Therefore, we encourage \sysname{} to generate captions that describe image regions relevant to the question instead of generic captions with no specific aim. 

We accomplish this goal by leveraging BLIP~\cite{li2022blip}, a large-scale pretrained vision-language model that contains a network branch outputting a similarity score $\text{sim}(v, t)$ between an image $v$ and a text $t$. This branch, called Image-grounded Text Encoder (ITE), employs a vision transformer~\cite{dosovitskiy2020image} that encodes the image, and a textual encoder that attends to the image features using cross-attention. As input to the image encoder, the image is equally divided into $K$ patches.
\input{table/fig_gradcam}

To identify relevant image patches, we feed the image $v$ and the question $t$ to the ITE network and apply a variation of GradCAM~\cite{selvaraju2017grad}, a feature-attribution interpretability technique, that aggregates all cross-attention maps using weights from the gradients. Formally, let us denote image patch features as $X \in \mathbb{R}^{K\times D_v}$, where $K$ is the number of image patches and $D_v$ the image feature dimension. We denote textual features as $Y \in \mathbb{R}^{M \times D_t}$, where $M$ is the number of textual tokens and $D_t$ the text feature dimension. For every cross-attention head, we have parameter matrices $W_Q\in \mathbb{R}^{D_t \times D_t}$ and $W_K\in \mathbb{R}^{D_v \times D_t}$ . The cross-attention scores, $A \in \mathbb{R}^{M \times K}$, can be written as
\begin{equation}
    A = \operatorname{softmax}\left(\frac{Y W_Q W_K^\top X^\top}{\sqrt{D_t}}\right).
\end{equation}
The $j$\textsuperscript{th} row of $A$ indicates the amount of attention the $j$\textsuperscript{th} textual token allocates to all image patches. 
At a selected layer of the ITE network, we compute the derivative of the similarity score w.r.t the cross-attention score, $\partial \text{ sim}(v, t) / \partial A$, and multiply the gradient matrix element-wise with the cross-attention scores. 
The relevance of the $i$\textsuperscript{th} image patch, $\text{rel}(i)$, takes the average over $H$ attention heads and the sum over $M$ textual tokens: 
\begin{equation}
    \text{rel}(i) = \frac{1}{H} \sum^{M}_{j=1} \sum^{H}_{h=1} \max\left(0,\frac{\partial  \text{ sim}(v, t)}{\partial A_{ji}^{(h)}}\right) A_{ji}^{(h)},
\end{equation}
where the superscript $^{(h)}$ denotes the index of attention heads. 
For every caption we generate, we sample a subset of $K^\prime$ image patches with probability proportional to the patch relevance. The captioning module sees the sampled patches only.  

We provide the following motivation for the technique. The attention matrix $A$ may be taken as indicative of patch importance. However, much redundancy exists among these matrices and many attention heads may be pruned with little performance loss \cite{bian-etal-2021-attention}, suggesting that some scores are uninformative. Inspired by GradCAM, we filter out uninformative attention scores by multiplication with the gradient which could cause an increase in the image-text similarity.

\input{table/tbl_patch_selection}

\input{table/fig_fusion}
Figure~\ref{fig:gradcam} shows some examples of generic captions and question-guided captions with associated relevance heatmaps.
We can clearly observe that question-guided captions contain more relevant information that helps produce the correct answers.

Table~\ref{tbl:patch_selection} gives a quantitative analysis about the effect of different patch selection methods on zero-shot VQA performance across three datasets.
Question-guided patch sampling substantially outperforms generic captioning using all patches and random patch sampling,
especially when the number of captions is large. 100 question-guided captions  outperform the 5 human-written captions from MS COCO by 5.2\% on VQAv2 and 6.0\% on OK-VQA, demonstrating the merit of the proposed approach.

\subsection{Informative Image Captioning}
\label{sec:informative-image-captioning}
Even with relevant image regions, there may still be more than one way to describe these regions. Some descriptions may contain the desired answer to the question, whereas others may not. Without the ability to identify the answer \emph{a priori}, we aim to generate maximally diverse captions to provide coverage of possible answers. 

We adopt the image captioning network branch from BLIP~\cite{li2022blip} and apply stochastic top-$k$ sampling~\cite{fan2018hierarchical} instead of beam search, which is known to produce dull and repetitive captions \cite{Vijayakumar:diverse-beam-search:2016,holtzman2019curious}. The input to the network contains the $K^\prime$ image patches sampled according to relevance (see \S \ref{sec:image-question-matching}).  We prepend a short prompt, ``a picture of '' as input to the text decoder. We repeat this process to generate $N$ captions per image to encourage diversity of captions and coverage of visual content. 
To prevent repetition, we keep a generated caption only if it is not subsumed by any previous caption as an exact substring.

\subsection{Answering the Question}
\input{table/fig_acc_fusion_enc_dec}
The question-answering encoder-decoder model is pretrained on text data only and can only process text. Therefore, we include the question and the generated captions as input to the model. As discussed in \S \ref{sec:informative-image-captioning}, the image captioning module generates multiple diverse captions. To process such long inputs efficiently, we adopt the Fusion-in-Decoder (FiD) strategy~\cite{fusion_decoder:izacard2020leveraging}. 

We illustrate the FiD strategy in Figure~\ref{fig:fusion} by comparing it with the more straightforward Fusion-in-Encoder (FiE), which concatenates the question and all captions into a long paragraph as input to the encoder. In contrast, FiD encodes each caption with the question separately and concatenates the \emph{encoded representations} of all tokens from all captions. The result is fed as input to the decoder and is processed  through the cross-attention mechanism. Since the time complexity of the self-attention mechanism scales quadratically with input length, whereas the cross-attention scales linearly with the encoder's output length, FiD is much more efficient than FiE. 
Further,
FiE is constrained by the maximum input length of the encoder, caused by the positional encoding, but FiD does not have this constraint. Hence, with FiD, \sysname{} can benefit from even more captions.

We plot the performance of FiD and FiE against the number of captions in Figure~\ref{fig:acc_fusion_enc_dec}. Initially, both methods improve as the number of captions increases. However, the performance of FiE is capped at around 40 captions when the maximum input length is exceeded, whereas the performance of FiD continues to rise.

%% file: table/fig_gradcam.tex
\begin{figure*}[!t]
	\begin{center}
	\includegraphics[width=1\linewidth]{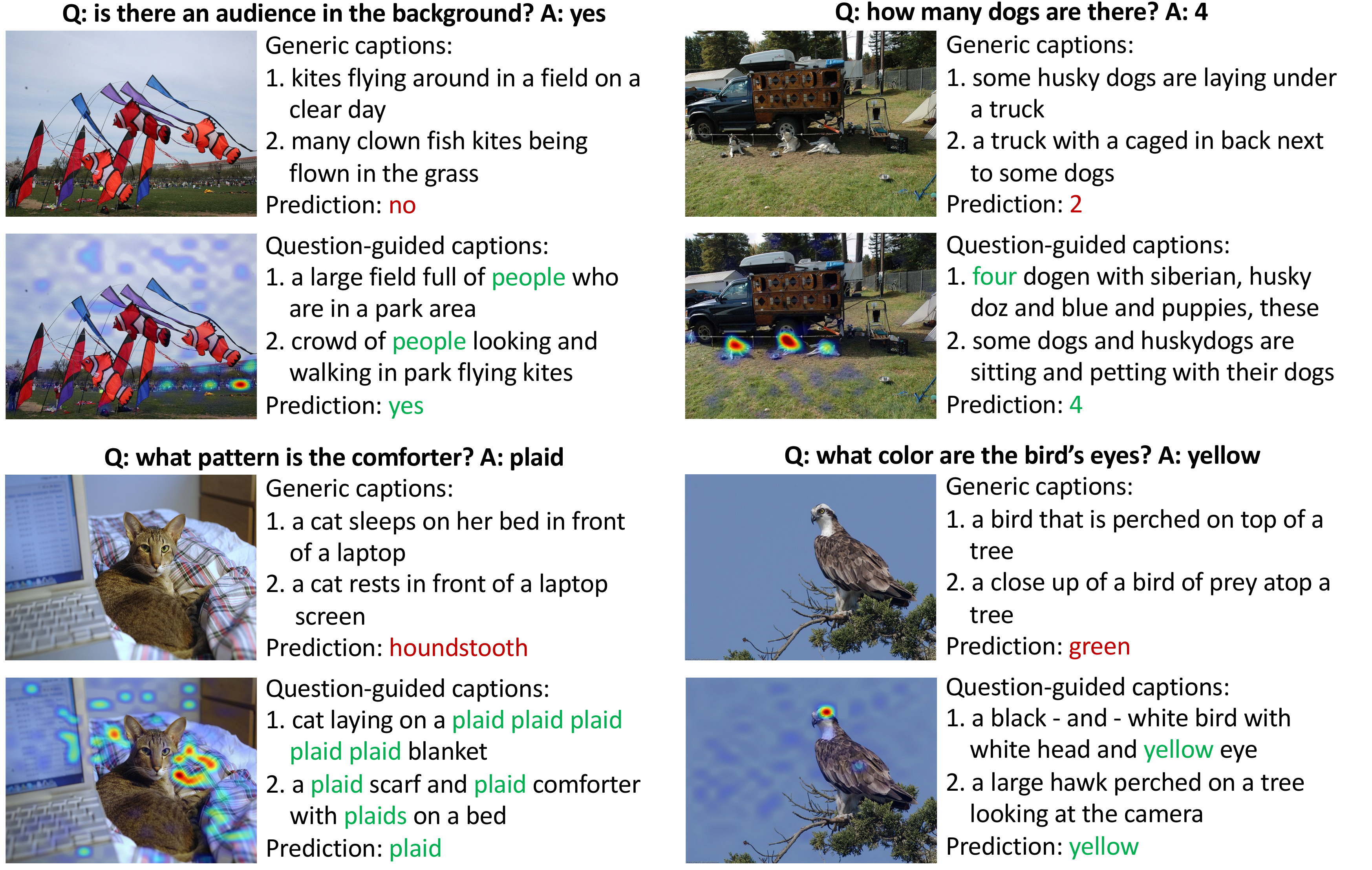}
	\vspace{-4ex}
	\end{center}
	\caption
	{
	Examples of generic captions (from all patches) based on the original image and question-guided captions (from the sampled patches) based on the GradCAM heatmaps on VQAv2 data. For illustrative purposes, we highlight words in green to indicate correct answer predictions and the cues from captions. Words in red indicate wrong answer predictions. 
	}
	\label{fig:gradcam}
\end{figure*}

%% file: table/tbl_patch_selection.tex
\begin{table*}[!t]
	
	\centering
	\resizebox{0.87\textwidth}{!}{
	\begin{tabular}	{l | c | c c c c}
	\hline	 	 	
			Image Patch Sampling Strategy & Num. of Captions & VQAv2 & OK-VQA & GQA  \\
	\hline
			\textit{No captions} & 0 & 33.4 & 10.3 &  25.9 \\
	\hline	 	 	
			All patches (generic captions)  & 5 & 53.5            & 26.6            & 36.5       \\
			Uniform random sampling  & 5 & 52.0             & 25.5            & 36.2     \\
			Question-guided patch sampling & 5 & 56.3 &  27.0 & 37.9 \\
			Human-written captions from MS COCO & 5 & 56.9        & 28.1  & -       \\	
	\hline
			All patches (generic captions) & 100 & 58.6       & 31.9          & 39.8    \\
			Uniform random sampling & 100 & 58.4        & 32.4           & 40.4     \\
			Question-guided patch sampling & 100 & \textbf{62.1}            & \textbf{34.1}     & \textbf{42.3}      \\
	\hline

	\end{tabular}
 }

	\caption
		{
		Comparison of different sampling strategies for image patches. 100 question-guided captions surpass the performance of 5 human-written captions from MS COCO. 
		}
	\label{tbl:patch_selection}

\end{table*}

%% file: table/fig_fusion.tex
\begin{figure*}[!t]

	\centering
	\includegraphics[width=0.8\textwidth]{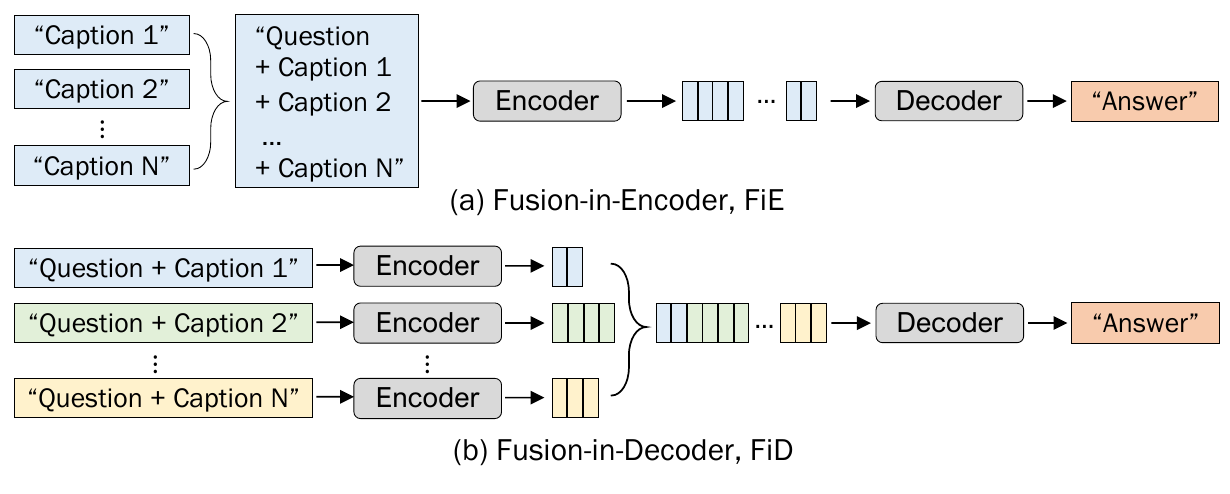}
    \vspace{-2ex}
  \caption
  	{
  	Two methods to process multiple captions with a question answering model. (a) Fusion-in-Encoder (FiE), which  concatenates the captions as a long input paragraph to the encoder. (b) Fusion-in-Decoder (FiD), which encodes each caption with the question individually and concatenates all encoded representations as input to the cross-attention mechanism of the decoder.
	  } 
  \label{fig:fusion}
 \end{figure*} 

%% file: table/fig_acc_fusion_enc_dec.tex
\begin{figure*}[!t]
	\centering
	\includegraphics[width=\textwidth]{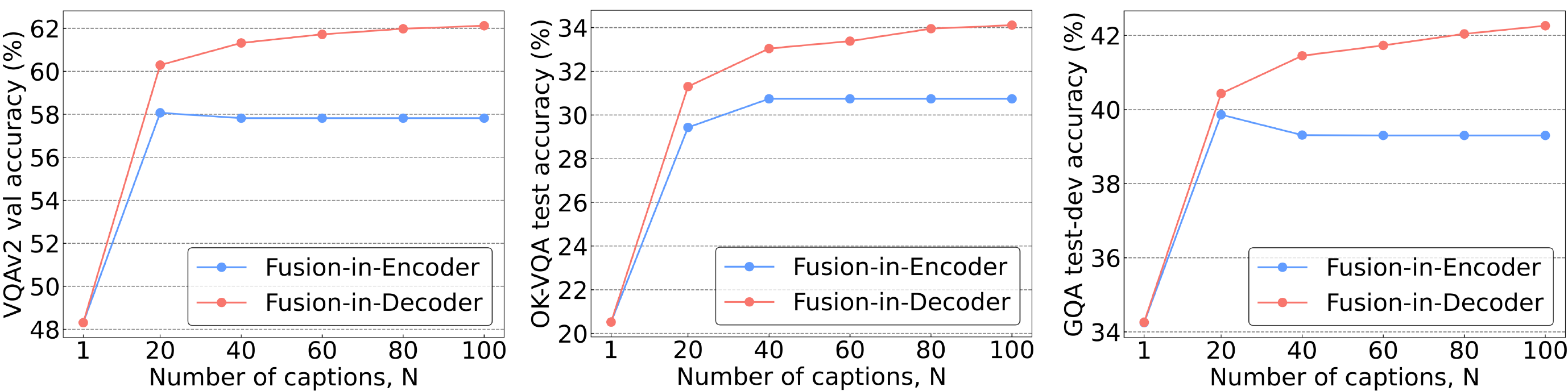}
  \caption
  	{
  		Comparison between Fusion-in-Encoder and Fusion-in-Decoder for VQAv2, OK-VQA and GQA. 
	  } 
  \label{fig:acc_fusion_enc_dec}
 \end{figure*} 

%% file: sec_experiment.tex
\section{Experiments}
\label{sec:sota}
\subsection{Datasets and Evaluation}

We adopt multiple zero-shot VQA benchmarks, including the validation set (214,354 questions) and test-dev set (107,394 questions) of VQAv2~\cite{goyal2017making}, the test set (5,046 questions) of OK-VQA~\cite{marino2019ok}, and the test-dev set (12,578 questions) of GQA-balanced~\cite{hudson2019gqa}.
We include the VQAv2 validation set as a few recent works~\cite{frozen:tsimpoukelli2021multimodal, fewvlm:jin2021good} evaluate their performance on this dataset only. 
We obtain the answer by open-ended generation and perform evaluation based on exact matching. 
We report soft-accuracy~\cite{goyal2017making} for VQAv2 and OK-VQA to account for multiple ground truth answer; for GQA, we report the standard accuracy.

\subsection{Implementation Details}
To obtain the image-question matching module and image captioning module, we adopt BLIP~\cite{li2022blip} with the ViT-L/16 architecture pretrained on 129M image-text pairs. The original BLIP-ITM and BLIP-Caption models further finetune on the 2017 train split of COCO Captions~\cite{coco}, which partially overlaps with VQAv2 and OKVQA. To prevent data leak, we instead finetune on the 2014 train split of COCO Captions, which does not overlap with the VQA evaluation datasets. We emphasize that this represents less, not more, training compared to the publicly released BLIP. 

For the question answering module, we adopt UnifiedQAv2~\cite{khashabi2022unifiedqa} trained on diverse textual QA datasets. It is worth noting that UnifiedQAv2 is completely unaware of the visual modality during training. Therefore, its training data do not overlap with the VQA datasets. 

Unless otherwise stated, we utilize a total of 100 captions per question.
We select the 8\textsuperscript{th} cross-attention layer of the ITE network for GradCAM. We sample $K'=20$ image patches for the generation of each caption,
and use $k=50$ for top-$k$ decoding (see Fig.~\ref{fig:num_patch_gradcam_layer} in Appendix B).
For VQAv2 and OK-VQA, we apply FiD and encode the question with one caption at a time. However, for GQA, we encode each question with a group of 5 captions.
GQA requires compositional visual reasoning and thus benefits from more contextual information per question. 
We perform experiments using LAVIS~\cite{lavis} on 8 Nvidia A100 GPUs. 
\input{table/tbl_sota}

\subsection{Comparison with State of the Arts}
We compare with state-of-the-art methods that formulate zero-shot VQA as open-ended answer generation.
We categorize the methods based on how the pretrained networks are conjoined. In the first group, including VL-T5\textsubscript{no-vqa}~\cite{cho2021unifying}, 
FewVLM~\cite{fewvlm:jin2021good},  VLKD~\cite{dai2022enabling}, Flamingo~\cite{Deepmind:Flamingo2022}, and Frozen~\cite{frozen:tsimpoukelli2021multimodal}, a vision encoder (VE) embeds the image as a dense matrix and feeds it to the pretrained language model (PLM). After that, the system performs a round of end-to-end vision-language (VL) training on tasks other than VQA, such as image captioning. VL-T5\textsubscript{no-vqa} and FewVLM freeze the VE and finetune the PLM, whereas Frozen freezes the PLM and trains the VE. VLKD finetunes both the PLM and part of VE. Flamingo partially finetunes both the VE and the PLM.
In the second group, the two foundation models are not jointly trained. Instead, they use language in the form of captions as the intermediate representation for an image. This group includes PICa~\cite{pica:yang2021empirical} and our proposed model, \sysname{}.

Table~\ref{tbl:sota} shows the results. 
\sysname{} outperforms previous methods by large margins on VQAv2 and GQA. 
On VQAv2 test-dev, \sysname{}\textsubscript{11B} outperforms the second best technique, Flamingo\textsubscript{80B}~\cite{Deepmind:Flamingo2022}, by 8.5\%. \sysname{}\textsubscript{3B} outperforms Flamingo\textsubscript{80B} by 7.2\% despite its significantly smaller size and the similar-sized Flamingo\textsubscript{3B} by 14.3\%.
On GQA, \sysname{}\textsubscript{large} outperforms the FewVLM\textsubscript{large} by 9.1\%, with similar-sized PLM despite the lack of end-to-end training. 
Only on OK-VQA, Flamingo performs better than \sysname{}. 
OK-VQA requires external knowledge not existing in the images and cannot be solved by good captions alone.
We hypothesize that the end-to-end training on the gigantic vision-language dataset of Flamingo induces a mapping between images and knowledge concepts that helps with OK-VQA.
However, \sysname{} is still better on OK-VQA than all other baselines that not trained on the gigantic Flamingo data. 
Compared with language-conjoined PICa~\cite{pica:yang2021empirical} with 175B parameters, \sysname{}\textsubscript{11B} achieves a sizable improvement of 18.2\%.

The results underscore the difficulty of zero-shot VQA using language models without any vision-language (VL) training. PICa, with its 175B-parameter language model, achieves comparable performance as FewVLM\textsubscript{large}, whose language model is 236x smaller but finetuned on VL data.
On the other hand, finetuning the billion-scale language model could incur heavy computational cost and risk catastrophic forgetting~\cite{frozen:tsimpoukelli2021multimodal, Deepmind:Flamingo2022}.
\sysname{} demonstrates the feasibility of a different paradigm: using billion-scale pretrained language models for VQA with zero training.

\section{Analysis}

\subsection{Are \sysname{} captions informative?}
\input{table/fig_cor_cap_ans_acc}

Intuitively, if the captions contain the correct answer,
the QA model would have a higher chance to answer correctly.
To measure the utility of captions,
we compute the \textit{answer hit rate} (AHR),
or the proportion of questions for which at least one caption contains the ground-truth answer verbatim. 
Here we exclude questions with yes/no answers as the meaning of ``yes'' and ``no'' can be contextual and these two words appear rarely in captions. 

\input{table/tbl_decode}

Figure~\ref{fig:cor_cap_ans_acc}(a) shows the correlation between the AHR and VQA accuracy, computed over the VQAv2 validation set, for three techniques of image patch sampling: question-guided sampling, uniform random sampling, and all patches. We observe that, within each sampling method, the VQA accuracy increases as the AHR increases. 
This corroborates our hypothesis that the presence of the answer in the captions facilitates the generation of the correct answer.

The correlation between performance and AHR is not perfect, as AHR does not capture other factors that may affect the answer accuracy, such as the position of the answer in the sentence and the number of its occurrence.
However, AHR provides an easy-to-compute and useful measure for the information quality of the captions.  

Figure~\ref{fig:cor_cap_ans_acc}(b) shows how AHR changes with the number of captions. Among the three techniques, question-guided sampling produces captions with the highest AHR. Thus, we may attribute the good performance of \sysname{} partially to its informative, question-guided captions that directly contain the correct answer. Further, as the number of captions increases from 20 to 100, question-guided AHR increases from 71.8\% to 84.0\%. This demonstrates the benefit of Fusion-in-Decoder, which allows \sysname{} to utilize up to 100 captions. 

\subsection{How sensitive is \sysname{} to the caption decoding method?}
\input{table/tbl_lang_transformer}

As the content of captions plays a crucial role in the performance of \sysname{}, we investigate the sensitivity to the choice of the caption decoding methods. We test four methods, including the deterministic beam search and three stochastic methods --- temperature sampling~\cite{ficler2017controlling, caccia2018language}, nucleus sampling~\cite{holtzman2019curious}, and top-$k$ sampling~\cite{fan2018hierarchical}. 
We generate 100 captions from each method, and report the results in Table~\ref{tbl:decode}.
\sysname{} performs very similarly across stochastic decoding methods, but beam search results in a noticeable drop. Upon close inspection, we observe that beam search generates repetitive captions that do not sufficiently cover different aspects of the image.

\subsection{Can \sysname{} work with other textual QA models?}
\label{subsec:language_transformer}
We experiment with two other PLMs as the question answering module for \sysname{}: T0~\cite{sanh2022multitask:T0} and GPT-J~\cite{gpt-j}.
T0 is an encoder-decoder model which is pretrained in a multi-task fashion 
on a collection of NLP tasks, including question answering. 
GPT-J is a decoder-only model, a much smaller open-source alternative to GPT-3~\cite{GPT-3}, 
which is pretrained with a task-agnostic language modeling loss on a large-scale text corpus.
Table~\ref{tbl:lang_transformer} shows that UnifiedQAv2 performs better on VQA tasks compared to T0 and GPT-J.
We attribute UnifiedQAv2's good performance to the fact that it is a task-specific question answering model with superior textual QA performance.
The result indicates that the choice of PLM is important when performing zero-shot VQA with zero training.
The modular and flexible design of \sysname{} leaves room for 
further performance improvements as more advanced PLMs emerge.

%% file: table/tbl_sota.tex
\begin{table*}[!tbp]
	 \vspace{-1ex}
	\centering
	\setlength\tabcolsep{3pt}
	\resizebox{1\textwidth}{!}{
	\begin{tabular}	{l|l c c |l c c | c c c c}
	\hline	
			\multirow{2}{*}{Method} & \multicolumn{3}{c|}{Language}  & \multicolumn{3}{c|}{Vision}  & \multicolumn{2}{c}{VQAv2} & OK-VQA & GQA  \\
			& Model & \#Params & VL-aware & Model & \#Params & VL-aware & Val & Test-dev & Test & Test-dev \\
	\hline	
	\multicolumn{11}{c} {\emph{Pretrained models conjoined by end-to-end VL training.}} \\[1ex]

			VL-T5\textsubscript{no-vqa} 
			& T5 & 224M & \checkmark & Faster R-CNN & 64M & \xmark & 13.5 & -  & 5.8 & 6.3  \\

			FewVLM\textsubscript{base}
			& T5 & 224M & \checkmark & Faster R-CNN & 64M & \xmark  & 43.4 & - & 11.6 & 27.0 \\ 
			FewVLM\textsubscript{large}
			& T5 &740M & \checkmark & Faster R-CNN & 64M & \xmark & 47.7 & - & 16.5 & 29.3 \\ 
            VLKD\textsubscript{ViT-B/16}
			& BART & 407M & \checkmark & ViT-B/16 & 87M & \checkmark  & 38.6 & 39.7 & 10.5 & - \\
			VLKD\textsubscript{ViT-L/14}
			& BART & 408M & \checkmark & ViT-L/14 & 305M & \checkmark & 42.6 & 44.5 & 13.3 & - \\
			Flamingo\textsubscript{3B} & Chinchilla-like & 2.6B & \checkmark & NFNet-F6 & 629M & \checkmark  & - & 49.2 & 41.2 & -  \\
			Flamingo\textsubscript{9B} & Chinchilla-like & 8.7B & \checkmark & NFNet-F6 & 629M & \checkmark  & - & 51.8 & \underline{44.7} & -  \\		
			Flamingo\textsubscript{80B} & Chinchilla & 80B & \checkmark & NFNet-F6 & 629M & \checkmark  & - & 56.3 & \textbf{50.6} & -  \\	
			Frozen
			& GPT-like & 7B & \xmark & NF-ResNet-50 & 40M &\checkmark & 29.5 & - & 5.9 & -  \\
	\hline	
	\multicolumn{11}{c} {\emph{Pretrained models conjoined by natural language and zero training. }} \\ [1ex]
        	PICa
        	& GPT-3 &  175B &  \xmark & VinVL-Caption &  259M & \checkmark & - & - & 17.7 & - \\

	        \sysname{}\textsubscript{base} & UnifiedQAv2 & 223M & \xmark & BLIP-Caption & 446M & \checkmark & 54.3 & 55.2 & 23.0 & 34.6 \\
	        \sysname{}\textsubscript{large} & UnifiedQAv2 & 738M & \xmark & BLIP-Caption & 446M & \checkmark & 57.5 & 58.8 & 27.1 & 38.4 \\ 
			\sysname{}\textsubscript{3B} & UnifiedQAv2 & 2.9B & \xmark & BLIP-Caption & 446M & \checkmark & \underline{62.1} & \underline{63.5} & 34.1 & \textbf{42.3} \\ 
			\sysname{}\textsubscript{11B} & UnifiedQAv2 & 11.3B & \xmark & BLIP-Caption & 446M & \checkmark & \textbf{63.3}  & \textbf{64.8} &  35.9  & \underline{41.9}   \\ 

	\hline		 	
	\end{tabular}
}
    \vspace{-1ex}
	\caption
		{
			Comparison with state-of-the-art models on zero-shot VQA. Flamingo~\cite{Deepmind:Flamingo2022} inserts additional parameters into the language model and perform training using billion-scale vision-language data. The best accuracy is bolded and the second best is underlined.
		}
	\label{tbl:sota}
\end{table*}

%% file: table/fig_cor_cap_ans_acc.tex
\begin{figure*}[!tbp]
	\centering
	\includegraphics[width=\textwidth]{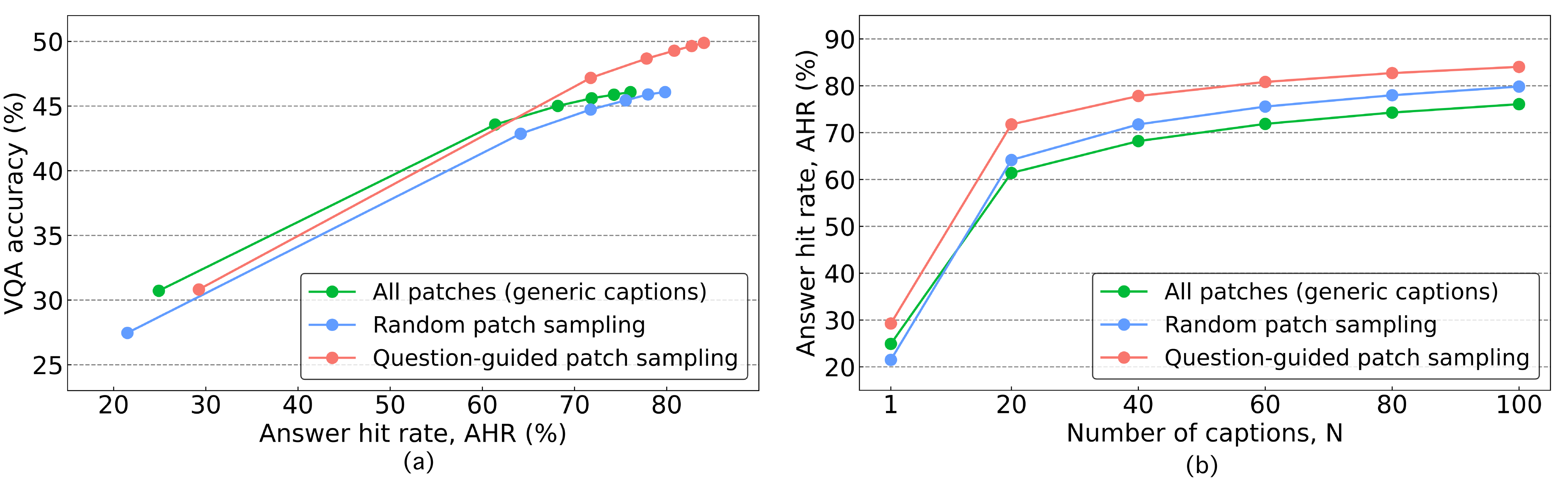}
    \vspace{-4ex}
  \caption
  	{
  	Analysis on the relationships between \textit{answer hit rate} (AHR),
  	VQA accuracy,
  	and the number of captions per question (N).
  	(a) shows a positive correlation between AHR and VQA accuracy.
  	(b) shows the AHR increases with N,
  	  	where the proposed question-guided patch sampling produces captions with the highest AHR.
	  } 
  \label{fig:cor_cap_ans_acc}

 \end{figure*}

%% file: table/tbl_decode.tex
\begin{table}[!t]
	
	\centering
 	\resizebox{1\columnwidth}{!}{
    \setlength\tabcolsep{4pt}
	\begin{tabular}	{l|c c c c}
	\hline		 	
			Decoding Method & VQAv2 & OK-VQA & GQA  \\
	\hline	
			Beam search & 55.3 & 26.8 & 37.2 \\
			Temperature ($t$=0.5) & 61.2 & 32.1 & 41.4 \\
			Temperature ($t$=1)  & 60.0 & 31.6 & 41.6 \\
			Nucleus ($p$=0.9) & 61.3 & 32.9 & 41.7 \\
			Nucleus ($p$=0.95) &  60.7 & 32.2 & 41.9 \\
            Top-$k$ ($k$=50) & \textbf{62.1} & \textbf{34.1} & 42.3 \\
            Top-$k$ ($k$=100) & 61.9 & 34.0 & \textbf{42.3} \\
	\hline	
	\end{tabular}
 }

	\caption
		{
			Ablation study on different caption decoding methods. 
			\sysname{}\textsubscript{3B} performs well across the stochastic methods. 
		}
	\label{tbl:decode}
\end{table}

%% file: table/tbl_lang_transformer.tex
\begin{table}[!t]
	\centering
 	\resizebox{1\columnwidth}{!}{
 	\setlength\tabcolsep{4pt}
	\begin{tabular}	{l c | c c c c}
	\hline	 	 	
			QA Model & \#Params & VQAv2 & OK-VQA & GQA \\
	\hline	
			GPT-J  & 6B & 28.7 & 14.5 & 18.5 \\ 

            T0 & 3B  & 49.6 & 26.6 & 32.3 \\
            T0 & 11B  & 47.3 & 30.5 & 33.4 \\
            
            UnifiedQAv2 & 3B  &  62.1 & 34.1 & \textbf{42.3} \\
            UnifiedQAv2 & 11B  &  \textbf{63.3} & \textbf{35.9} & 41.9 \\
	\hline	
	\end{tabular}
}

	\caption
		{
				Ablation study on various textual question answering module for \sysname{} on zero-shot VQA. UnifiedQAv2 is a task-specific model pretrained for question answering.
		}
	\label{tbl:lang_transformer}
\end{table}

%% file: sec_conclusion.tex
\section{Conclusion}
\label{sec:conclusion}
We propose \sysname{}, a framework with zero additional training for zero-shot VQA by conjoining off-the-shelf pretrained models.
\sysname{} leverages an image-question matching module to determine image patches relevant to the current question.
An image captioning module then generates question-guided captions, which are processed by a question answering module to produce an answer.
\sysname{} achieves state-of-the-arts performance on multiple VQA benchmarks.
We hope that our work will bring inspiration for further research in flexible, modular AI systems for solving vision-language tasks.

\section{Limitations}
\label{sec:limitations}
Like two sides of the same coin, the strengths and weaknesses of \sysname{} both result from the zero-training modular system design.
\sysname{} enjoys the power of pretrained models but also inherits the bias from these models.
It enjoys the efficiency of zero training,
but introduces additional inference cost due to the multi-step process.
Nevertheless,
we believe that the strengths of \sysname{} outweigh its limitations,
and welcome further investigations to help debias pretrained models and improve inference speed.

\section{Acknowledgments}
Anthony Meng Huat Tiong is supported by Salesforce and Singapore Economic Development Board under the Industrial Postgraduate Programme. 
Boyang Li is supported by the Nanyang Associate Professorship and the National Research Foundation Fellowship (NRF-NRFF13-2021-0006), Singapore. Any opinions, findings, conclusions, or recommendations expressed in this material are those of the authors and do not reflect the views of 
the funding agencies.


%% file: sec_appendix.tex
\section{Visualization}
In the appendix, we show visualizations of GradCAM heatmaps and the generated captions for VQAv2, OK-VQA, and GQA in following pages.  
\input{table/fig_appendix_gradcam_vqa}
\input{table/fig_appendix_gradcam_okvqa}
\input{table/fig_appendix_gradcam_gqa}
\input{table/fig_appendix_itm_patch_gradcam}

\section{Hyperparameter sensitivity}

\label{sec:appendix}

We study how VQAv2 validation accuracy varies with different cross-attention layer 
used for GradCAM and number of image patches sampled for question-guided caption generation. Figure~\ref{fig:num_patch_gradcam_layer}(a) shows no clear relationship between VQA accuracy and the cross-attention layer used for GradCAM. 
The maximum difference in VQA accuracy across different cross-attention layers is 3\%. 
Figure~\ref{fig:num_patch_gradcam_layer}(b) shows that VQA accuracy has a negative correlation with the number of sampled image patches. As $K'$ increases, the sampled patches become less relevant to the questions,
and question-guided patch sampling becomes akin to using all patches.

%% file: table/fig_appendix_gradcam_vqa.tex
\begin{figure*}[!htbp]
	\begin{center}
	\includegraphics[width=\linewidth]{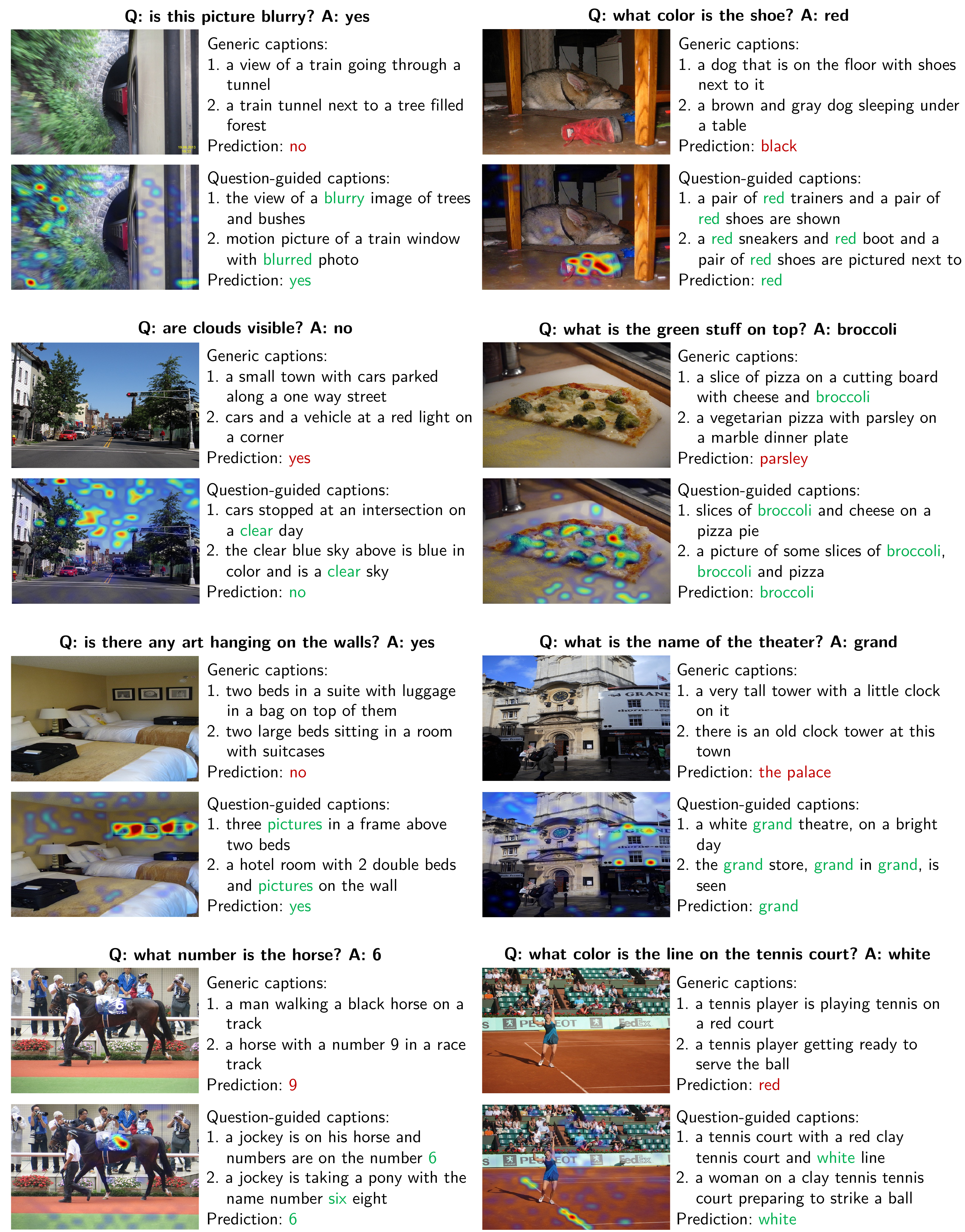}
	\vspace{-4ex}
	\end{center}
	\caption
	{
	Examples from \textbf{VQAv2}. We show generic captions (from all patches) based on the original image and question-guided captions (from the sampled patches) based on the GradCAM heatmaps. For illustrative purposes, we highlight words in green to indicate correct answer predictions and the cues in captions. Words in red indicate wrong answer predictions. 
	}
	\label{fig:gradcam_vqa_appendix}
\end{figure*}

%% file: table/fig_appendix_gradcam_okvqa.tex
\begin{figure*}[!t]
	\begin{center}
	\includegraphics[width=1\linewidth]{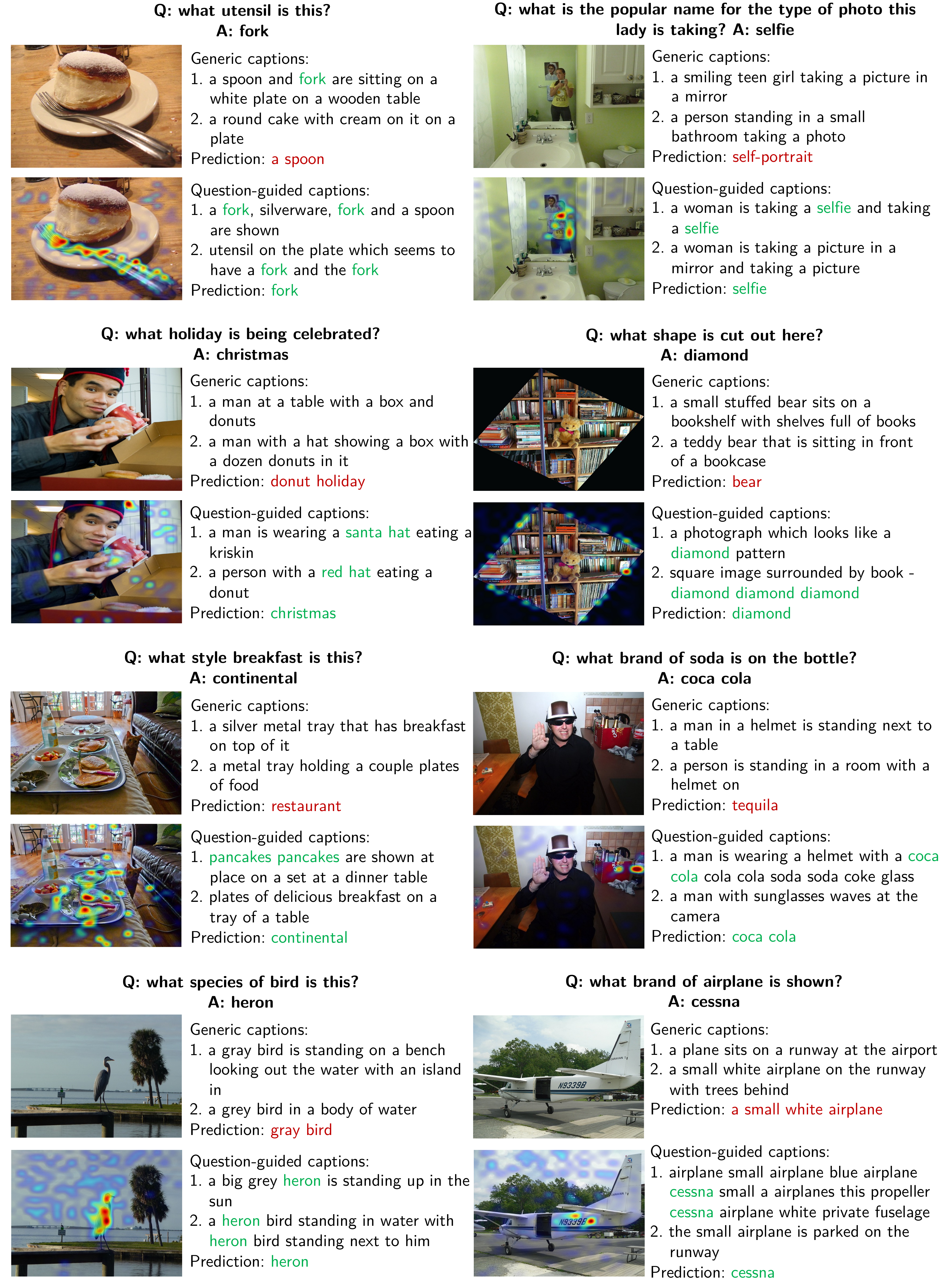}
	\vspace{-5ex}
	\end{center}
	\caption
	{
	Examples from \textbf{OK-VQA}. We show generic captions (from all patches) based on the original image and question-guided captions (from the sampled patches) based on the GradCAM heatmaps. For illustrative purposes, we highlight words in green to indicate correct answer predictions and the cues in captions. Words in red indicate wrong answer predictions. 
	}
	\label{fig:gradcam_okvqa_appendix}
\end{figure*}

%% file: table/fig_appendix_gradcam_gqa.tex
\begin{figure*}[!t]
	\begin{center}
	\includegraphics[width=1\linewidth]{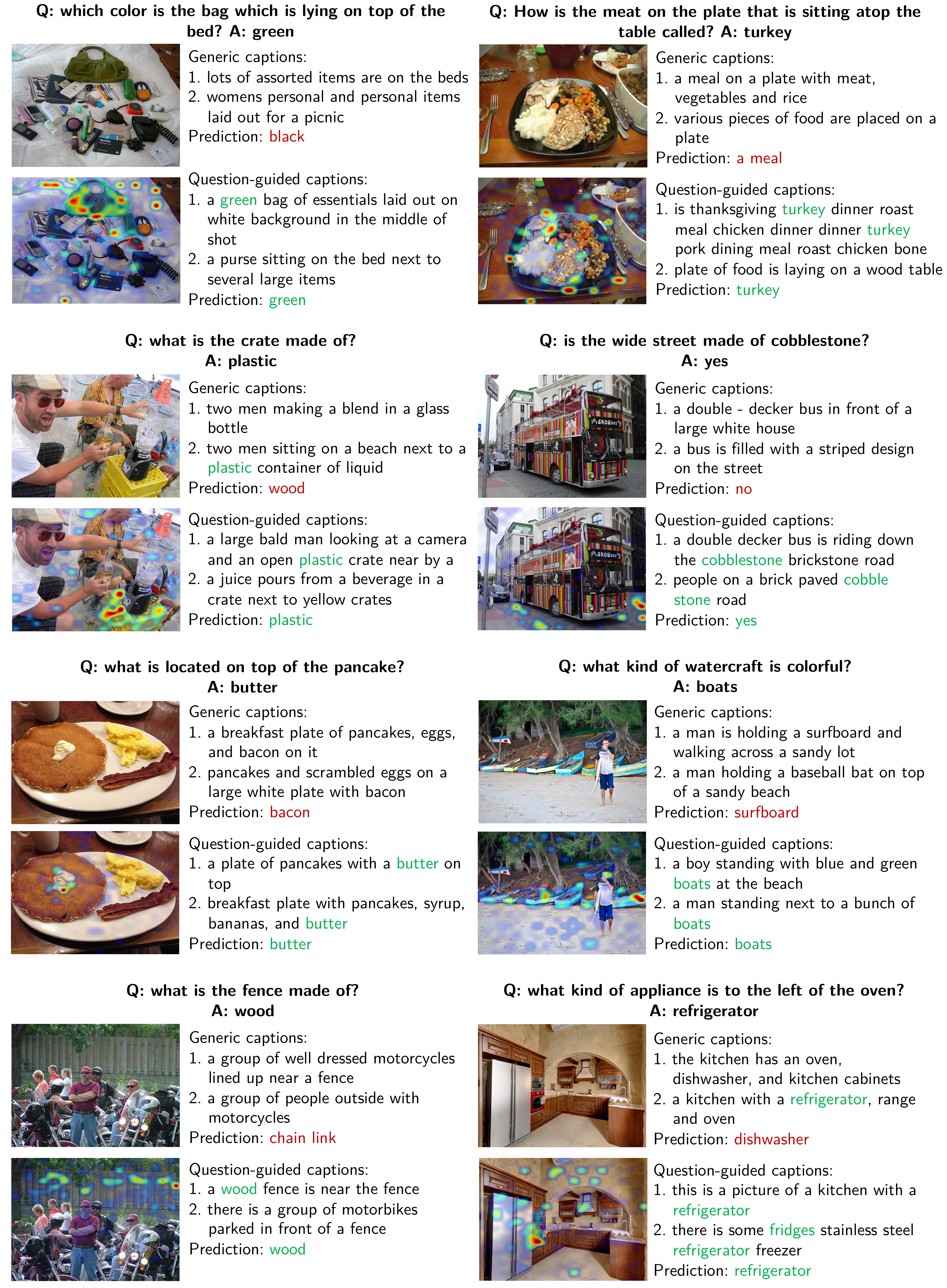}
	\vspace{-4ex}
	\end{center}
	\caption
	{
	Examples from \textbf{GQA}. We show generic captions (from all patches) based on the original image and question-guided captions (from the sampled patches) based on the GradCAM heatmaps. For illustrative purposes, we highlight words in green to indicate correct answer predictions and the cues in captions. Words in red indicate wrong answer predictions. 
	}
	\label{fig:gradcam_gqa_appendix}
\end{figure*}

%% file: table/fig_appendix_itm_patch_gradcam.tex
\begin{figure*}[!t]
	\centering
	\includegraphics[width=\textwidth]{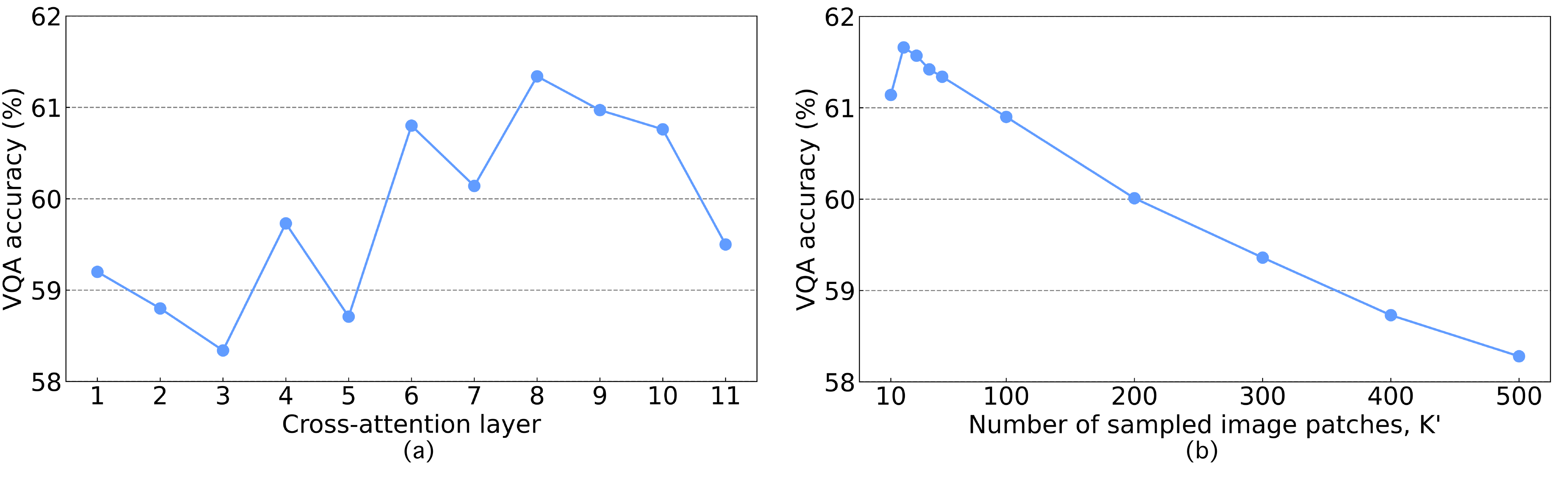}
  \caption
  	{
  		VQAv2 validation set accuracy using (a) different cross-attention layer on which GradCAM is computed using $K'= 50$. (b) different number of image patches sampled for caption generation using GradCAM computed at 8\textsuperscript{th} cross-attention layer.
	  } 
  \label{fig:num_patch_gradcam_layer}
 \end{figure*}